\documentclass[journal]{IEEEtran}
\usepackage{times}
\usepackage{epsfig}
\usepackage{graphicx}
\usepackage{amsmath, amsfonts, amssymb}

\usepackage{url}            % simple URL typesetting
\usepackage{booktabs}       % professional-quality tables
\usepackage{nicefrac}       % compact symbols for 1/2, etc.
\usepackage{microtype}      % microtypography
\usepackage{algorithm}
\usepackage{algpseudocode}
\usepackage{graphicx}
\usepackage{amsmath}
\usepackage{multirow}
\usepackage{color}
\usepackage{subfigure}
\usepackage{tabularx}

\ifCLASSINFOpdf
\else
\fi
\begin{document}
\title{Separable Batch Normalization for Robust Facial Landmark Localization with Cross-protocol Network Training}

\author{Shuangping~Jin,
        Zhenhua~Feng,
        Wankou~Yang,
        Josef~Kittler
%        and~Jane~Doe,~\IEEEmembership{Life~Fellow,~IEEE}% <-this % stops a space
%\thanks{M. Shell was with the Department
%of Electrical and Computer Engineering, Georgia Institute of Technology, Atlanta,
}

\maketitle

\begin{abstract}
A big, diverse and balanced training data is the key to the success of deep neural network training. However, existing publicly available datasets used in facial landmark localization are usually much smaller than those for other computer vision tasks. A small dataset without diverse and balanced training samples cannot support the training of a deep network effectively. To address the above issues, this paper presents a novel Separable Batch Normalization (SepBN) module with a Cross-protocol Network Training (CNT) strategy for robust facial landmark localization.

Different from the standard BN layer that uses all the training data to calculate a single set of parameters, SepBN considers that the samples of a training dataset may belong to different sub-domains. Accordingly, the proposed SepBN module uses multiple sets of parameters, each corresponding to a specific sub-domain. However, the selection of an appropriate branch in the inference stage remains a challenging task because the sub-domain of a test sample is unknown. To mitigate this difficulty, we propose a novel attention mechanism that assigns different weights to each branch for automatic selection in an effective style. As a further innovation, the proposed CNT strategy trains a network using multiple datasets having different facial landmark annotation systems, boosting the performance and enhancing the generalization capacity of the trained network. The experimental results obtained on several well-known datasets demonstrate the effectiveness of the proposed method.
\end{abstract}

% \begin{IEEEkeywords}
% , IEEEtran, journal, \LaTeX, paper, template.
% \end{IEEEkeywords}

% \fi
\IEEEpeerreviewmaketitle

\begin{figure*}[!t]
	\centering
	\includegraphics[width=.9\linewidth]{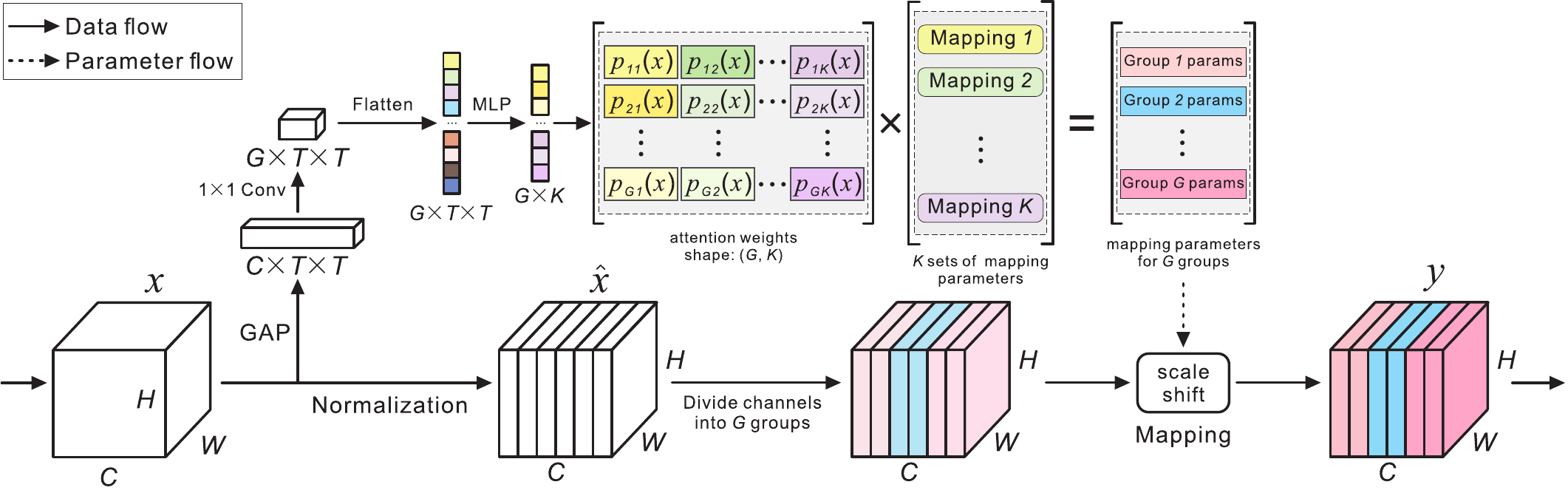}
	\caption{The proposed SepBN module with $K$ sets of parameters. The input feature map $\mathcal{X}$ is normalized as the standard BN layer to get $\hat{\mathcal{X}}$ which then divided into $G$ groups. The multiple sets of mapping parameters are weighted and summed using the attention weights to produce group-specific parameters for the final mapping operation from $\hat{\mathcal{X}}$ to $\mathcal{Y}$.}
	\label{fig:sepbn}
	\vspace{-1em}
\end{figure*}

\section{Introduction}
\IEEEPARstart{F}{acial} landmark localization aims to predict the position of a set of pre-defined facial key points.
It plays a crucial role in many automatic face analysis systems, including face recognition~\cite{taigman2014deepface, liu2017sphereface, deng2019arcface}, face frontalisation~\cite{hassner2015effective}, emotion estimation~\cite{zeng2008survey, walecki2016copula, li2017reliable}, 3D face fitting~\cite{zhu2017face, guo2020towards}, etc.
The rapid development of this research area during the past years produced a variety of effective neural network architecture designs~\cite{wu2018look, yang2017stacked} and loss functions~\cite{feng2018wing, teixeira2019adaloss, wang2019adaptive}, which have been instrumental in achieving impressive results.
The reported performance of these deep Convolutional Neural Network~(CNN-) based methods demonstrates their superiority over traditional approaches such as the Active Shape Model~(ASM)~\cite{cootes1995active}, Active Appearance Model~(AAM)~\cite{cootes2001active} and cascaded-regression-based methods~\cite{wu2016constrained, wu2017simultaneous, feng2017face} especially when handling faces in the wild.

In unconstrained scenarios, the key challenge for facial landmark localization is posed by facial appearance variations, including pose, illumination, expression, occlusion, motion blur, low image resolution etc. All these influencing factors should be taken into account but this is impeded by the difficulty to collect samples with all different appearance variations.
The existing datasets in the facial landmark localization community are usually much smaller than those available for other computer vision tasks.
For example, the AFLW~\cite{burgos2013robust} dataset has only 20K training samples, but it is almost the largest training set for facial landmark localization. 
A worse case is the COFW~\cite{burgos2013robust} dataset that has only 1345 training samples. 
Such a small dataset leads to the inevitable data imbalance issue across different sub-domains in facial landmark localization.
As a result, the trained network may not be able to generalize well for unseen samples whose types occur rarely in the training set. 

In a unique way, we try to deal with the aforementioned issues by performing Separable Batch Normalization~(SepBN) in deep neural network training. 
The classical BN layer has two operations, normalization and mapping.
These two operations of BN simply force the mean/variance of a tensor across specific dimensions to be 0/1 and further scale and shift the normalized tensor.
However, the classical BN layer treats all the samples of a dataset equally, leading to potential bias and overfitting of a trained network.
In contrast, SepBN alleviates this issue by integrating $K$ separable BN branches in parallel. 
These branches share the same normalization operation but maintain $K$ different sets of mapping parameters. Then,
$G$ groups of mapping parameters are produced via a novel attention mechanism that only depends on the input feature map to scale and shift the corresponding $G$ groups of channels as shown in Figure~\ref{fig:sepbn}.
The SepBN module endows the original BN layer with a non-linear mapping capability to map the normalized feature map dynamically using the information obtained by the attention block. 
More importantly, the proposed SepBN module addresses the small-sample-size issue of an unbalanced training data effectively.

To further address the aforementioned issues, we advocate a Cross-protocol Network Training~(CNT) strategy that consolidates multiple datasets with different landmark annotation systems, \textit{e.g.}, AFLW with 19 landmarks and WFLW~\cite{wu2018look} with 98 landmarks, by means of a multi-task learning manner for performance boosting. 
Thanks to this problem formulation, the proposed CNT strategy is able to extract commonly used facial features that improve the performance of the trained network on individual facial landmark dataset.
In addition, this training strategy also mitigates the data imbalance issue and boosts the performance of the trained network on a training dataset with a small number of samples.

The main contributions of the proposed method are:
\begin{itemize}
	\item A novel SepBN module that is able to deal with the unbalanced data issue in deep network training for facial landmark localization. Specifically, the training samples from different sub-domains are normalized and mapped through different branches automatically using an attention fusion scheme.
	
	% 	\item An attention mechanism that is able to automatically learn the group information 
	
	\item An effective cross-protocol network training strategy that utilizes multiple facial landmark localization datasets with different landmark annotation protocols for network training, moderating the unbalanced data distribution problem further and alleviating the risk of over-fitting on small datasets. 
	
	\item A comprehensive validation of the proposed SepBN module and CNT strategy on both simple and advanced network architectures, including our Vanilla CNN, ResNeXt~\cite{xie2017aggregated} and MobileNet~\cite{sandler2018mobilenetv2}. The results obtained on several benchmarking datasets demonstrate the effectiveness of the proposed method.
\end{itemize}

\section{Related work}
\label{sec_2}
\textbf{Facial landmark localization } has been studied for decades, resulting in a variety of well-known approaches, from the traditional methods like ASM~\cite{cootes1995active}, AAM~\cite{cootes2001active} and their variants~\cite{romdhani1999multi, hu2004fitting} to modern deep-learning-based methods~\cite{zhang2016joint, yang2017stacked, wu2018look, kumar2019uglli, wang2019adaptive, kumar2020luvli}.

Two main deep-learning-based facial landmark localization methods have been developed: heatmap-based methods~\cite{yang2017stacked, wu2018look, wang2019adaptive} and coordinate-regression-based methods~\cite{feng2018wing, feng2019mining}. 
The above two types of methods take a facial image as input while output the facial landmarks in two different ways. 
A heatmap-based method outputs landmarks using 2D heatmaps, in which the pair of coordinates with the highest response value corresponds to the landmark position.
In contrast, a coordinate-regression-based method solves the landmarking problem by predicting the landmark coordinates directly.

The heatmap-based facial landmark localization methods exploit the beneficial properties of the U-Net like networks~\cite{ronneberger2015u}, such as the Hourglass network~\cite{newell2016stacked} and densely connected U-Nets (DU-Net)~\cite{tang2019towards}. 
These kinds of networks are characterized by their bottom-up and top-down architecture and their maintenance of continuous relational interconnection, organized in an encoder-decoder fashion. 
However, heatmap-based methods  suffer from the problem of quantization errors.
Additionally, the training of such a network involves more hyper-parameters and its success often depends on making use of special tricks. 
Recently, a new function, Soft-Argmax~\cite{luvizon2019human}, has been proposed to integrate the two mainstreams by viewing heatmaps as probability distributions and calculating the expectations of the distributions as landmark locations.
This method can be directly supervised by facial landmark coordinates.

In our work, we focus on the coordinate-regression-based method and propose a new  SepBN module to enhance the network learning capability.
It should be highlighted that, to the best of our knowledge, this is the first time that a new BN module has been developed for the facial landmark localization task. Most of the existing related methods are designed for classification tasks but there are significant differences between the two tasks.

The well-known \textbf{Batch Normalization} method alleviates the convergence problems in deep neural network training effectively~\cite{ioffe2015batch}.
BN normalizes the features by mean and variance in the ($N$, $H$, $W$) dimensions of the feature map along $C$ channels, where $N$ is the batch size, and $H, W$ are the spatial resolution of one feature map channel. 

In recent years, more advanced BN methods have been proposed, such as layer normalization~\cite{ba2016layer}, instance normalization~\cite{ulyanov2016instance}, group normalization~\cite{wu2018group}, and others~\cite{shao2019ssn, ortiz2020local, yao2020cross}. 
Commonly these studies aim to improve the performance of the trained network by changing some aspects of the  normalization process.
For example, the layer normalization method focuses on each individual sample by normalizing the features in the ($C$, $H$, $W$) dimension~\cite{ba2016layer}. 
Accordingly, each training sample is normalized by its own data, thus obviating  the issue of instability in parameter estimation because the normalization operation becomes uncorrelated with batch size $N$. 
The instance normalization method normalizes a feature map along the ($H$, $W$) dimensions~\cite{ulyanov2016instance}.
It focuses on the mean and variance of each channel of a sample. 
The mean and variance in instance normalization are calculated with fine granularity, resulting in good performance in low-level tasks, such as style transferring. 
Group normalization is a trade-off between layer normalization and instance normalization. It normalizes the features in the ($C^\prime$, $H$, $W$) dimensions where $C^\prime$ is smaller than $C$~\cite{wu2018group}. 
Group normalization provides more stable performance than the classical BN layer, showing robustness to the variation of batch size.
\begin{figure}[!t]
	\centering
	\includegraphics[width=.81 \linewidth]{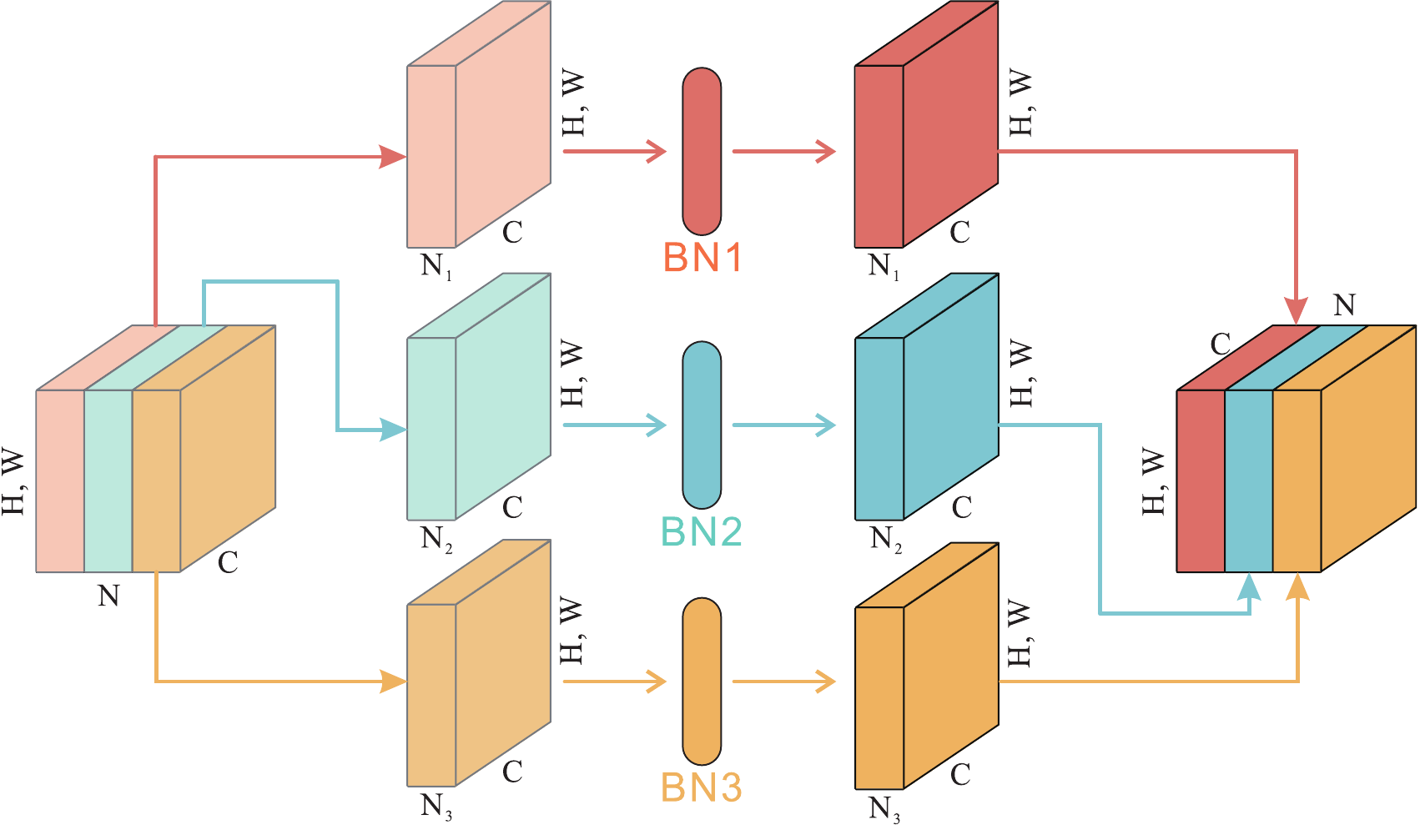}
	\caption{A brute-force SepBN module with 3 separate routes.}
	\label{fig:sepBN}
	%\vspace{-1em}
\end{figure}

In general, existing normalization layers share the same design pattern: \textit{normalization} and \textit{mapping}. 
Nevertheless, the role of the mapping operation has been little investigated in the literature. 
Our proposed SepBN module normalizes the features in the ($N$, $H$, $W$) dimension just like the standard BN layer. 
The key difference between SepBN and other normalization modules lies in the \textit{mapping} operation as shown in Figure~\ref{fig:sepbn}. 
Recently, a new normalization method called attentive normalization~\cite{li2019attentive} was proposed to integrate feature normalization and feature attention together, however, as we will show in the experiment, AN provides limited performance boost in facial landmark localization. By the way, SepBN could be used in place of AN as a more general and powerful module. 

% \textbf{Attention mechanism: }Attention mechanism first appeared in the field of natural language processing in [AttentionIsAllYouNeed], and was later introduced into the field of computer vision by SENet. In recent years, more and more sophisticated attention modules have been developed for better performance while some of them can only be used under restricted circumstances because of their high complexity. 

% Without bells and whistles, the attention mechanism embedded in our proposed SepBN module is lightweight and intuitive, resulting in its efficiency and interpretability and making it widely applicable.

%The goal is to minimize the gap between the predicted coordinates vector $\mathbf{v}_{pred}$ and the ground-truth coordinates vector $\mathbf{v}_{gt}$ on training set as much as possible and at the same time ensure the generalization ability of the learned function. The commonly used L1 loss function is applied to supervise the learning process.

\section{Separable Batch Normalization}
\label{sec_3}
In this section, we first introduce the brute-force SepBN module that directly forces the training samples from different distribution domains to pass through different BN branches, as shown in Figure~\ref{fig:sepBN}. Based on the analysis of the learned parameters in the brute-force SepBN module, we find that the \textit{tracking} parameters in standard BN have a limited influence on the final performance whereas the \textit{mapping} parameters play a more important role. Based on this finding, the normalization operation of the brute-force SepBN module are merged to estimate shared \textit{tracking} parameters and retain the separate \textit{mapping} for different domains. Furthermore, to reduce the reliance on face data priors and enhance the mapping ability, we introduce a novel attention mechanism to assign different weighted sums of mapping parameters to different channel groups without any prior information, as shown in Figure~\ref{fig:sepbn}.

\subsection{Brute-force SepBN}
The standard BN layer consists of two key parts: 
\begin{equation}
\hat{\mathcal{X}}_c = \frac{\mathcal{X}_c-\mathrm{E}(\mathcal{X}_c)}{\sqrt{\operatorname{Var}(\mathcal{X}_c)}},
\mathcal{Y}_c = \gamma_c \hat{\mathcal{X}}_c + \beta_c,
\end{equation}
where $\mathcal{X}_c \in \mathbb{R}^{N \times H \times W}$ is one patch of the input tensor $\mathcal{X} \in \mathbb{R}^{N \times C \times H \times W}$, $\mathrm{E}()$ and $\operatorname{Var}()$ calculates the mean and variance within the patch. $\gamma_c$ and $\beta_c$ further scale and shift the normalized tensors. To be more specific, the first equation normalizes the mean and variance of each patch to be $0$ and $1$ to reduce internal covariate shift.  The running mean and variance of the corresponding patch will be updated in a moving average manner according to the calculated value. In order to preserve the linearity of BN, the normalized patch is scaled and shifted by $\gamma_c$ and $\beta_c$ in the second equation. There are two key components in a BN layer: \textit{tracking} parameters~(running mean and variance, updated in the forward phrase) and \textit{mapping} parameters~(scale and shift, updated by network backward propagation) which correspond to the normalization and mapping operations, respectively. 
\begin{figure}[!t]
	\centering
	\includegraphics[trim = 0mm 0mm 0mm 2mm, clip, width=.93\linewidth]{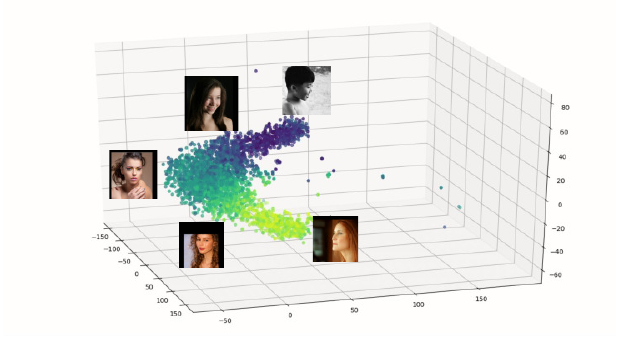}
	\caption{The t-SNE visualization of the AFLW training set~\cite{burgos2013robust}. All the faces are roughly distributed into three sub-domains.}
	\label{fig:tsne}
	% 	\vspace{-1em}
\end{figure}

As a statistically significant module, BN is normally applied regardless of the data domain origin, that is to say, data is always processed in the same way even if they belong to different domains. It seems that the existing facial landmark localization methods ignore this fact and use one single BN to track the mean and variance of all the data and map them across a mini-batch. However, in practice, face samples are naturally drawn from different distributions. For example, Figure~\ref{fig:tsne} shows the distribution of the AFLW dataset~\cite{burgos2013robust}. We can see that 
all the faces are roughly clustered in three domains conditioned by head pose. %\textcolor{red}{\textbf{Please complete the above few sentences.}}

To address this issue, we first design a brute-force SepBN module so as to deal with data from different domains separately. 
As demonstrated in~Figure~\ref{fig:sepBN}, different BN layers are used in parallel to track the mean and variance of the data from specific domains and normalize/map them separately. Specifically, data from different domains are concatenated along the batch dimension. When passing through a brute-force SepBN module, the feature maps from the same domain are gathered and directed to pass through a BN branch specifically set up for this domain. Consequently, the running mean and variance values of different domains can be tracked separately. Hence, potential unhelpful interactions among domains are prevented and inaccurate estimation of tracking parameters is mitigated. The mapping parameters are domain-specific as well. 

\begin{table}[!t]
	\centering
	\footnotesize
	\caption{The configuration of Vanilla CNN. We set the negative slope of LeakyReLU as $10^{-2}$, kernel size and stride of Max Pooling as 2.} 
	\label{vanillacnn}
	\begin{tabular}{ll}\toprule
		Input & Operator           \\
		\midrule
		$128 \times 128 \times 3$ & Conv($3 \times 3 \times 64$), BN, LReLU, MAX \\
		$64 \times 64 \times 64$  & Conv($3 \times 3 \times 128$), BN, LReLU, MAX \\
		$32 \times 32 \times 128$ & Conv($3 \times 3 \times 256$), BN, LReLU, MAX \\
		$16 \times 16 \times 256$ & Conv($3 \times 3 \times 512$), BN, LReLU, MAX \\
		$8 \times 8 \times 512$   & Conv($3 \times 3 \times 1024$), BN, LReLU, MAX \\
		$4 \times 4 \times 1024$  & Conv($3 \times 3 \times 2048$), BN, LReLU, MAX \\
		$2 \times 2 \times 2048$  & Flatten \\
		8192  & Linear(8192, 1024), LReLU \\
		1024  & Linear(1024, $2L$) \\
		\bottomrule                         
	\end{tabular}
\end{table}

\begin{table}[!t]
	\footnotesize
	\renewcommand{\arraystretch}{1}
	\centering
	\caption{A comparison of the classical BN and the proposed brute-force SepBN in terms of the Normalized Mean Error~(NME) on the AFLW dataset, using the Vanilla CNN model.}
	\footnotesize
	\label{tableattentionfree}
	\begin{tabular}{lc}
		\toprule
		Method & NME \\ 
		\midrule
		%CCL~\cite{zhu2016unconstrained} & 2.72$\times10^{-2}$ \\
		%SAN~\cite{dong2018style} & 1.91$\times10^{-2}$ \\
		%DSRN~\cite{miao2018direct} & 1.86$\times10^{-2}$ \\
		Vanilla CNN (BN) & 1.65$\times10^{-2}$  \\
		Vanilla CNN (brute-force SepBN) & 1.56$\times10^{-2}$  \\
		\bottomrule
	\end{tabular}
	\vspace{-1em}
\end{table}

To verify the effectiveness of the brute-force SepBN module, we train two Vanilla CNN networks equipped with the classical BN and our brute-force SepBN on the AFLW dataset~\cite{burgos2013robust}. 
The architecture of the Vanilla CNN network is presented in Table~\ref{vanillacnn}. The input of our Vanilla CNN is an RGB image $\mathcal{I}$ of size $128 \times 128 \times 3$. The Vanilla CNN predicts the landmark coordinate vector $\mathbf{v}_{pred} = [x_1, y_1, x_2, y_2, \dots, x_L, y_L]^T \in \mathbb{R}^{2L}$ directly, where $L$ is the number of landmarks.
We use 3 separate routes for our brute-force SepBN by splitting the AFLW training set into 3 subsets: near-frontal, left profile and right profile face sets, based on the t-SNE visualization result shown in Figure~\ref{fig:tsne}. 
The training samples of a specific subset will only go through the corresponding BN branch in the brute-force SepBN module. 
The final results are shown in Table~\ref{tableattentionfree}. 
We can see that the proposed brute-force SepBN module improves the performance of the Vanilla CNN network equipped with the standard BN layer significantly.
It should be noted that the testing of the Vanilla CNN network equipped with our brute-force SepBN is not straightforward because the domain of a test image is unknown, and thus the appropriate branch cannot be determined.
Therefore, all the test images are processed three times and the best regression result is kept to calculate the localization error.
The results only demonstrate the potential of the proposed brute-force SepBN module.
We will address this issue in the next section by using an attention-based fusion method.

\begin{figure}[!t]
	\centering
	\includegraphics[trim = 1cm 0cm 1cm 1cm, clip, width=1\linewidth]{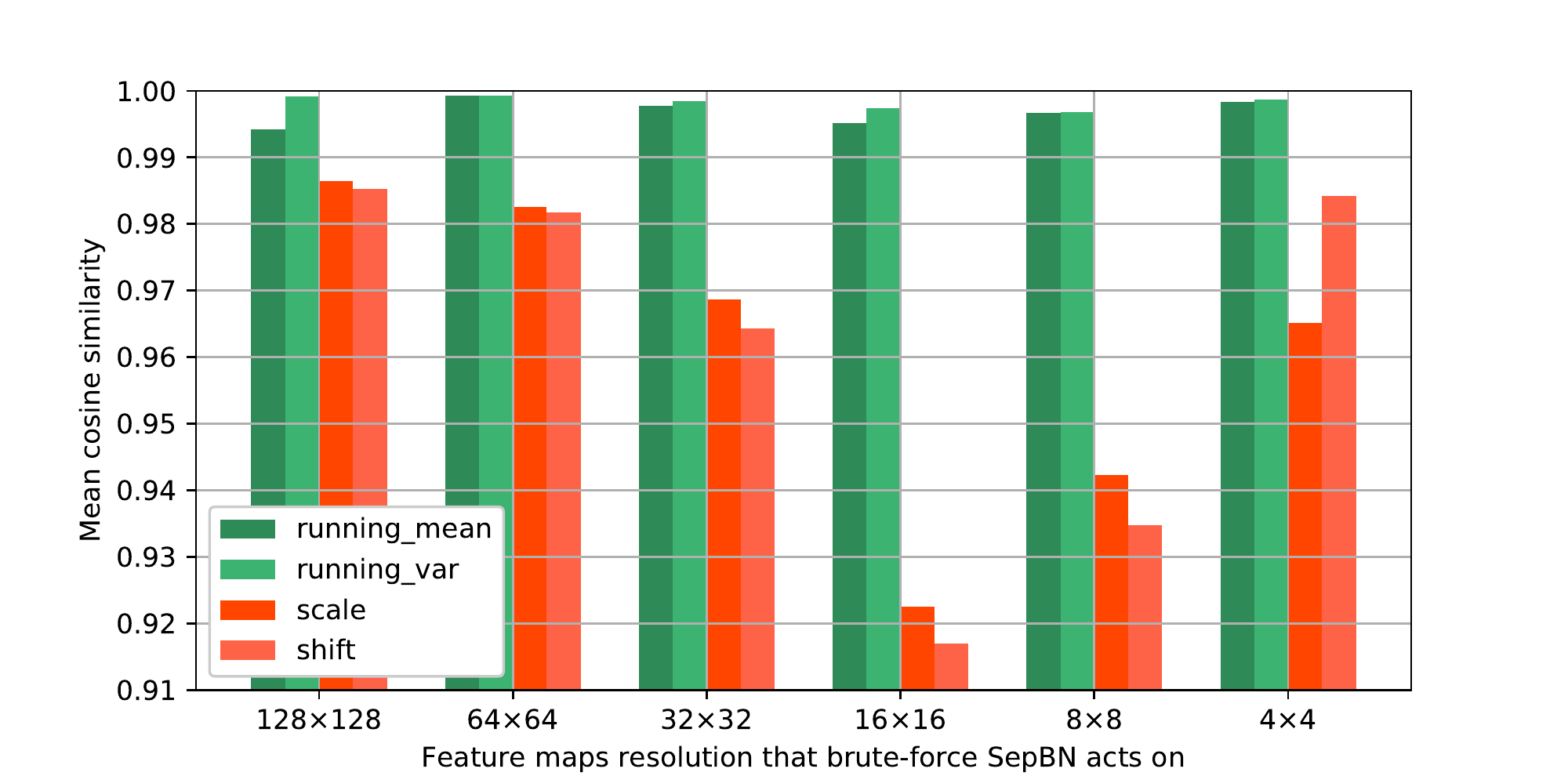}
	\caption{Mean cosine similarity of the running mean, running var, scale and shift learned in the brute-force SepBN modules across the network. Since our experiment is carried out by applying brute-force SepBN modules with 3 separate branches, the mean similarity is calculated as in Equation~\ref{equ3}.}
	\label{fig:bnsimilarity}
	\vspace{-1em}
\end{figure}

Note that there are two main procedures in a BN layer: normalization and mapping.
To further expose the component that leads to performance boosting, we calculate the average similarity of the learned tracking and mapping parameters of the three BN layers by:
\begin{equation}
\label{equ3}
S = [s(\mathbf{p}_{1},\mathbf{p}_{2}) + s(\mathbf{p}_{1}, \mathbf{p}_{3})+ s(\mathbf{p}_{2}, \mathbf{p}_{3})] / 3 
\end{equation}
where $s()$ calculates the cosine similarity of two given vectors, $\mathbf{p}_{k}$ are the learned tracking parameters (including running mean and variance) or mapping parameters (including scale and shift) of the $k$th branch.
Since we use 6 brute-force SepBN modules for the Vanilla CNN (1 for each convolution block) to replace the original 6 BN layers, $6 \times 4$ similarities can be calculated as shown in Figure~\ref{fig:bnsimilarity}. 
Surprisingly, we can see that the \textit{tracking} parameters of all three branches in any SepBN module are highly similar to each other throughout the whole network. 
In contrast, the learned \textit{mapping} parameters differ more significantly.
Based on the above observation, we conclude that the differences among mapping parameters are the key factors that boost the final performance of the trained network.

\subsection{Automatic SepBN with Attention}
\label{sectheproposedSepBN}
Based on the above discussions, the normalization operations of different BN layers embedded in a brute-force SepBN module can be merged while different sets of mapping operations need to be kept separate. 

Next, we focus on making the module choose the appropriate set of mapping parameters automatically and end-to-end trainable without any prior dataset division.
To be more specific, the proposed SepBN module maintains $K$ sets of mapping parameters $\gamma \in \mathbb{R}^{K \times C}$ and $\beta \in \mathbb{R}^{K \times C}$ and each of them shares the same size with that embedded in the standard BN layer. The purpose is to allow the network to select the most suitable mapping parameters for each sample. 

The simplest choice is applying a squeeze-and-excitation block like~\cite{hu2018squeeze} to produce attention weights $\lambda \in \mathbb{R}^{N \times K \times 1 \times 1}$ for the given $K$ sets of mapping parameters in the form of:
\begin{equation}
\operatorname{softmax} \left[ F_{ex} \left(F_{sq} \left(\mathcal{X}; \theta_{sq}\right); \theta_{ex}  \right) \right] = \lambda \in \mathbb{R}^{N \times K \times 1 \times 1},
\end{equation}
where $F_{sq}$ is the squeeze operation with the reduction rate $r$ involving a global average pooling layer, a linear transformation layer and a non-linear activation function, $F_{ex}$ is the excitation function involving a linear transformation layer and a sigmoid layer, the output of $F_{sq}$ and $F_{ex}$ are denoted as $\mathcal{X}_{sq} \in \mathbb{R}^{N \times \frac{C}{r} \times 1 \times 1}$ and $\mathcal{X}_{ex} \in \mathbb{R}^{N \times K \times 1 \times 1}$, $\theta_{sq}$ and $\theta_{ex}$ are the corresponding model parameters, $\operatorname{softmax}$ is an additional softmax function. 

Afterwards, the re-calibrated mapping parameters can be calculated by:
\begin{equation}
\hat{\gamma}_{n} = \sum_{k=1}^{K} \lambda_{n, k} \gamma_{k},~~
\hat{\beta}_{n} = \sum_{k=1}^{K} \lambda_{n, k} \beta_{k},
\end{equation}
where $\lambda_{n, k}$ is the attention weight for the $k$th set of mapping parameters (\textit{i.e.} $\gamma_{k}$ and $\beta_{k}$) of the $n$th sample, $\hat{\gamma}_{n}$ and $\hat{\beta}_{n}$ are the instance-specific mapping parameters used for the $n$th sample. 
Note that the attention block is applied to estimate the probability that a feature $\mathcal{X}$ needs to be mapped by the $k$th set of mapping parameters rather than acting directly on $\mathcal{X}$. 
By the way, the softmax function is used because we want to make the SepBN module behave in line with the standard BN layer in certain circumstances, \textit{i.e.} when the softmax function outputs a one-hot vector. 

Following~\cite{chen2020dynamic}, the temperature-controlled softmax function with an extra parameter $\tau$ is used as:
\begin{equation}
\lambda=\frac{\exp \left(\mathcal{X}_{ex} / \tau\right)}{\sum_{n} \exp \left(\mathcal{X}_{ex} / \tau\right)}.
\end{equation}
In this equation, the original softmax function can be considered as a special case of the temperature-controlled softmax function, \textit{i.e.} $\tau = 1$. 
The temperature parameter adjusts the sensitivity of the softmax function to the characteristics of the input data. 
By setting $\tau$ to a larger value, the output tends to a uniform distribution. 
According to~\cite{chen2020dynamic}, the temperature-based annealing training strategy converges better and provides better performance by setting a large temperature value for early epochs and gradually reducing it to 1 as the training progresses. 

However, the above simple SepBN ignores two important fact. First, attention weights are just used to assign different weights to different sets of parameters, which means only by applying a large enough $K$ can the module learn enough powerful mapping ability. Second, the simple SepBN makes each attention weight to act on one entire set of mapping parameters, and this is likely to bring about an embarrassing situation, that is, some valuable mapping parameters are likely to be inactivated due to the overall lower attention weight and vice versa. 

To solve the above mentioned drawbacks, the idea of channel grouping is introduced into our proposed SepBN module. Considering that the mapping parameters is to scale and shift each channel, each set of mapping parameters are divided into $G$ groups (channels of the normalized tensor $\hat{\mathcal{X}}$ will also be grouped in the same way) and then the attention block needs to generate attention weights $\pi \in \mathbb{R}^{N \times G \times K}$ of each group of channels in each set of mapping parameters for each sample. In order to effectively obtain such attention weights, a different attention mechanism is developed as Figure~\ref{fig:sepbn} shows. To be specific, the input feature $\mathcal{X} \in \mathbb{R}^{N \times C \times H \times W}$ is adaptively pooled (max pooling) into $\mathcal{X}_{amp} \in \mathbb{R}^{N \times C \times T \times T}$. The global average pooling is not chosen since we hope that more feature information can be retained by setting $T > 1$. Afterwards, $\mathcal{X}_{amp}$ will pass through a $1 \times 1$ convolution layer and output a new tensor in the shape of $N \times G \times T \times T$. Then the new tensor is flattened and used as the input of a Multi-Layer Perceptron module followed by a temperature-controlled softmax function. At last, using the generated attention weights, the appropriate scale and shift parameters for each group of channels of each sample can be calculated by, 
\begin{equation}
\begin{aligned}
\hat{\gamma}_{n, g} &= \sum_{k=1}^{K} \pi_{n, g, k} \gamma_{k},~~
\hat{\beta}_{n, g} = \sum_{k=1}^{K} \pi_{n, g, k} \beta_{k}, \\
&\text {s.t.} ~~ 0 \leq \pi_{n, g, k} \leq 1, \sum_{k=1}^{K} \pi_{n, g, k}=1 \\
\end{aligned}
\end{equation}
where $\pi_{n, g, k}$ is the attention weight indicating the probability that the $g$th channel group of the $n$th sample uses the $k$th set of mapping parameters (\textit{i.e.} $\gamma_{k}$ and $\beta_{k}$), $\hat{\gamma}_{n, g} \in \mathbb{R}^{M}$ and $\hat{\beta}_{n, g}  \in \mathbb{R}^{M}$ are the instance-and-group-specific mapping parameters used for the $g$th group of channels of the $n$th sample, $M$ is the channel number of each group. 
Then the mapping operation runs by,
\begin{equation}
\mathcal{Y}_{n, g} = \hat{\gamma}_{n, g} \hat{\mathcal{X}}_{n, g} + \hat{\beta}_{n, g}.
\end{equation}

Compared with the brute-force SepBN and simple SepBN, the proposed SepBN modification brings three major benefits. 
First, since the most proper mapping parameters can be learned automatically by using the attention mechanism, we do not have to split the training dataset into subsets corresponding to different feature domains manually. 
Second, the regressed landmark coordinates can be obtained directly by forwarding the input face images only once, without knowing the distribution information of a test sample in advance. 
Last but not least, through group division, rich combinations of mapping parameters can be obtained even when a smaller $K$ is used, making the module both lightweight and expressive.

\begin{figure}[!t]
	\centering
	\includegraphics[width=\linewidth]{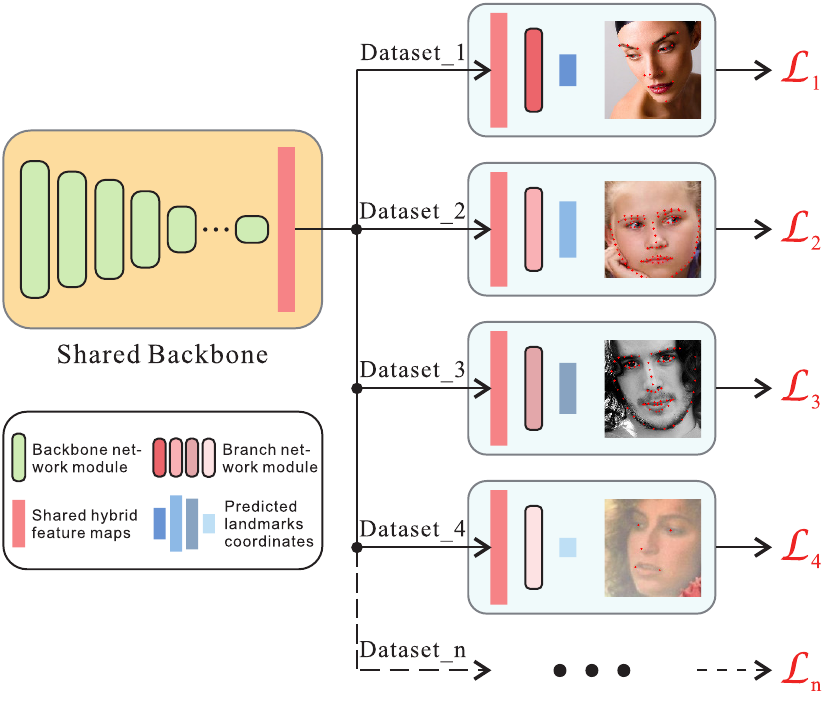}
	\caption{An overview of the proposed Cross-protocol Network Training (CNT) strategy. The shared backbone is used to extract common features that are helpful for different regression targets. Various branches are added after the backbone to regress different coordinate vectors. Note that, to make the shared feature maps adaptive to a specific landmark definition, an additional convolution layer is applied at the beginning of each branch.}
	%\textcolor{red}{(This figure should be re-plotted. This should be a generic figure.)}
	\label{fig:networkarchi}
	\vspace{-1em}
\end{figure}

\section{Cross-protocol Network Training}
Along with the advances in the methodology of facial landmark localization, many benchmarking datasets have been released to promote the development of the area, such as COFW~\cite{burgos2013robust}, AFLW~\cite{burgos2013robust}, WFLW~\cite{wu2018look} and 300W~\cite{sagonas2013300}. However, most of the existing 2D facial landmark localization datasets  have a common flaw, \textit{i.e.}, the amount of data is too small compared with that of other computer vision tasks such as object detection, image classification etc. Overall, the difference in the amount of data available is of two orders of magnitude. This is undoubtedly a huge problem for deep neural networks training that requires rich and diverse data. The hidden danger is that there is a high probability of overfitting and the problem of data imbalance will be magnified. 
In addition, the labeling schemes between different datasets are not standardized, making them hard to be jointly used no matter what kind of landmark regression method is applied, \textit{i.e.}, heatmap-based regression methods need to output heatmaps with different channel numbers and coordinate-regression-based methods need to output coordinates vectors with different length. In view of this, it can be concluded that a single facial landmark localization dataset is invariably non-uniformly distributed and of insufficient size. 

To mitigate this reality, and to make a full use of the existing datasets, as well as alleviating the data imbalance problem, we propose a Cross-protocol Network Training~(CNT) strategy comprising of two stages. 
In the first stage, a CNN network is trained with multiple supervision branches as illustrated in Figure~\ref{fig:networkarchi}. 
To be more specific, all the convolution layers constitute the shared backbone that is used to extract common facial features. 
Bearing in mind the various annotation schemes of different datasets, several branches emanate from the final shared feature maps to regress the coordinates vectors separately.
This makes sense because regardless of the number of landmarks involved in the regression and their semantics, the features that need to be extracted for localization are common. 
In the training process, the probability of sampling faces from each dataset is proportional to the size of the dataset and the sample will only select its corresponding branch for forwarding. 
The L1 loss is used for each branch for network training.
%Note that the well-known WIDERFACE face detection dataset also provides five facial landmarks on faces annotated by~\cite{deng2019retinaface} and this further expands our hybrid dataset.
Thus, all the existing facial landmark localization datasets can be utilized to form a richer, bigger and diverse dataset. 

The first stage of CNT provides the shared backbone with a more comprehensive feature extraction capability. 
However, the regression layers at the end of the network may not be fully trained especially for the branch corresponding to a dataset of small size.
Besides, the difference in the respective distributions of the datasets makes it necessary to perform adaptation to an individual feature domain.
Therefore, in the second stage of CNT, the network is fine-tuned to each dataset. 
In this stage, the redundant branches are disconnected and only the dataset-specific branch retained. 
Since we want to avoid destroying the comprehensive feature extraction capability learned by the shared backbone, the learning rate of the backbone is set to a relatively smaller value and that of the branch network is adjusted normally.
To the best of our knowledge, this is the first method for joint network training with multiple datasets for facial landmark localization.
Based on the experimental results reported in the next section, we can see that the proposed CNT method further boosts the performance of the trained network on each benchmarking dataset.

%%% Results
%%%------------------------------------------------------------
%%%------------------------------------------------------------
%%%------------------------------------------------------------
%%%------------------------------------------------------------
%%%------------------------------------------------------------
%%%------------------------------------------------------------
\section{Experiments}
\label{experiment}
\subsection{Implementation Details and Settings}

\textbf{Datasets:} We evaluated the proposed method on three datasets: COFW~\cite{burgos2013robust}, AFLW~\cite{koestinger2011annotated} and WFLW~\cite{wu2018look}.
The \textbf{COFW} dataset has 1,345 images for training and 507 images for test. 
COFW was designed to test the robustness of a facial landmark localization algorithm for faces with occlusions so most of the faces in COFW are occluded. 
Each face in COFW was manually annotated with 29 landmarks.
\textbf{AFLW} consists of 24,386 faces with large pose variations.
The AFLW dataset has two benchmarking protocols: AFLW-Full and AFLW-Frontal, both containing 20,000 training images. AFLW-Full uses all the remaining 4386 images for test but AFLW-Frontal only uses 1314 near-frontal faces for test. In AFLW, each face was annotated with 19 landmarks.
The \textbf{WFLW} dataset is a newly collected dataset that contains 10,000 faces (7,500 for training and 2,500 for test) with 98 facial landmarks. 
WFLW includes rich attribute annotations such as pose, illumination, make-up and so on.
The test set is divided into several subsets to verify the robustness of a algorithm for specific appearance variation types. 

\textbf{Evaluation Metric: } We used the Normalized Mean Error (NME) metric for evaluation, calculated as:
\begin{equation}
\mathrm{NME}(\%)=\frac{1}{L} \sum_{j=1}^{L}  \frac{\left\|\mathbf{p}_{j}-\mathbf{g}_{j}\right\|_{2}}{d} \times 100
\end{equation}
where $\mathbf{p}_{j}$ and $\mathbf{g}_{j}$ denote the $j$th predicted and ground-truth landmark locations, $d$ is the normalization term. 
For AFLW, we used the face box size as the normalization term.
For COFW and WFLW, we followed~\cite{wu2018look} and used the inter-ocular distance as the normalization term. 

\textbf{Implementation details}: All our experiments were implemented in PyTorch on a platform with one Intel Core i7-7700K CPU and one NVIDIA GeForce GTX 1080Ti GPU. We used the Stochastic Gradient Descent (SGD) optimizer and set the weight decay to 5e-4, momentum to 0.9 and batch size to 8. For learning rate, we used the Cosine annealing scheduler and set the maximum and minimum learning rate to $1\mathrm{e}-3$ and $5\mathrm{e}-6$, with 120 warm-up epochs. Each model was trained for 500 epochs with the L1 loss function.

We followed~\cite{feng2019mining} to perform data augmentation using the geometric augmentation methods. All the face images were cropped according to the official bounding box and resized to the size of $128\times 128$. We randomly rotated the training images between [-25, 25] and perturbed the bounding box corner within 15\% of the bounding box size. In addition, horizontal flip and shear transformations were applied. %For a fair comparison, all the test images were cropped and resized according to the bounding boxes provided without any transformation. 
The proposed SepBN module with $K=3$ separate routes was used to replace the classical BN layer. And $T=3$ and $G=2$ were found to be the most suitable setting for our task. The initial $\tau$ in the temperature-controlled softmax function of the attention mechanism was set to 30 and reduced to 1 in the first 30 epochs.

The datasets used in our Cross-protocal Network Training (CNT) strategy include all the training images of COFW, AFLW, WFLW, 300W and WiderFace~\cite{yang2016wider}. Specifically, for WiderFace, only the faces whose length and width of the bounding box are both greater than 80 are retained.
% Afterwards, the square hull that enclosed the 5 landmarks is extended by a quarter of the side length in each direction as the final bounding box to crop the face. 
The total number of the samples is 41763~(1345 from COFW, 20000 from AFLW, 7500 from WFLW, 3148 from 300W and 9770 from WiderFace), which is much larger than most of the existing facial landmark datasets. In the fine-tune phrase, the learning rate of the shared backbone is set to $1\mathrm{e}-7$ and that of the branch network is adjusted as described above.

% Besides, Cumulative Errors Distribution~(CED) curve is also used for evaluation.

% \begin{figure}
% 	\centering
% 	\includegraphics[width=\linewidth]{fig/AFLW_CED}
% 	\caption{A comparison of the CED curves on the AFLW dataset, including a set of state-of-the-art approaches:  RCPR~\cite{burgos2013robust}~, CFSS~\cite{zhu2015face}, LBF~\cite{ren2014face}~, GRF~\cite{hara2014growing}~, CCL~\cite{zhu2016unconstrained}~ and DAC-CSR~\cite{feng2017dynamic}.}
% 	\label{fig:aflwced}
% 	\vspace{-1em}
% \end{figure}
\renewcommand{\arraystretch}{1.1}
\begin{table}[!b]
	\scriptsize
	\centering
	\caption{Hard-aggregation vs. Soft-aggregation}
	\label{table:aggregation}
	\begin{tabular}{ccc}
		\toprule 
		& \multicolumn{2}{c}{ {NME(\%)} } \\
		& AFLW-Full & AFLW-frontal \\
		\midrule
		Vanilla CNN (BN) & $1.65$ & $1.40$ \\
		Vanilla CNN (SepBN, Hard-aggre) & $1.86$ & $1.60$ \\
		Vanilla CNN (SepBN, Soft-aggre) & $\mathbf{1.55}$ & $\mathbf{1.39}$  \\
		\bottomrule
	\end{tabular}
\end{table}

\subsection{Ablation study}
We performed a number of ablation studies on AFLW and WFLW using our Vanilla CNN to find the best configuration and validate the proposed method. For all experiments, we use the same hyper-parameters mentioned earlier except for the parameter that needs to be compared, and NME is used to show the effect of different settings in a clear way.

\textbf{Aggregation method: }There are two methods to obtain the final output of the SepBN module, hard aggregation (\textit{i.e.} $\hat{\gamma}_{n, g} = \arg\max_{k} \pi_{n, g, k} \gamma_{k},~
\hat{\beta}_{n, g} = \arg\max_{k} \pi_{n, g, k} \beta_{k}$) and soft aggregation (\textit{i.e.} $\hat{\gamma}_{n, g} = \sum_{k=1}^{K} \pi_{n, g, k} \gamma_{k},~
\hat{\beta}_{n, g} = \sum_{k=1}^{K} \pi_{n, g, k} \beta_{k}$). The results of these two different aggregation methods obtained on AFLW using our Vanilla CNN are shown in Table~\ref{table:aggregation}. 
We can see that the soft aggregation method performs much better. 
When hard-aggregation is applied, the parameters of the attention part cannot be effectively optimized, resulting in random attention weights, so the final performance is even worse than that of the baseline method using the classical BN layer. 

\textbf{Temperature annealing: } For SepBN, the hyper-parameter $\tau$ used in the temperature-controlled softmax function can be set to a fixed value or adjusted in the annealing way. 
We report the results obtained by different settings in Table~\ref{table:annealing}. 
Clearly, the annealing strategy outperforms all the other configurations.

\textbf{Number of separate routes $K$: } The number of sets of mapping parameters~(\textit{i.e.} $K$) may affect the performance of the trained network as well. 
We report the results of our Vanilla CNN trained with different values of $K$ in Table~\ref{table:routes}. We can see that, by setting $K$ to 3, we have the best result.
This empirically meets the visualization plot of the AFLW dataset in Figure~\ref{fig:tsne}.

\textbf{Number of channel groups $G$: } There are two ways to set the group number, one is fixing the channel number of each group and another one is fixing the group number. We selected the latter one to ensure that the module is sufficiently lightweight. And the experiments on AFLW shows that $G=2$ has already brought enough mapping ability for our task as shown in Table~\ref{table:groups}. 

\textbf{Number of adaptive pooling size $T$: }Adaptive max-pooling layer is applied in the attention block because it can not only make the network structure fixed, but also allow more input features to be retained. According to the experiment result in Table~\ref{table:poolingsize}, we empirically chose $T=3$ as the pooing size. 

\textbf{Effect in different depth of the network: }To find out how SepBN modules affect the performance across layers, we replaced the standard BN layers with the SepBN modules at different input resolutions and compared the performance on WFLW in Table~\ref{table:dyanmic-layers}. The results show that replacing all the original BN layers with SepBN achieves the best performance for our Vanilla CNN.

\begin{table}[t]
	\centering
	\scriptsize
	\caption{A comparison of different $\tau$(left) and $K$(right) configurations on the AFLW dataset using the Vanilla CNN network.}
	\begin{minipage}{0.4\linewidth}
		% 		\centering
		\label{table:annealing}
		\begin{tabular}{cc}
			\toprule
			Temperature &  NME(\%) \\
			\midrule
			$\tau = 1$ & $1.69$ \\
			$\tau = 5$ & $1.63$ \\
			$\tau = 10$ & $1.64$ \\
			$\tau = 20$ & $1.67$ \\
			$\tau = 30$ & $1.66$ \\
			Annealing $\tau$ & $\mathbf{1.55}$ \\
			\bottomrule
		\end{tabular}
	\end{minipage}
	\hspace{1em}
	\begin{minipage}{0.4\linewidth}
		% 		\centering
		\label{table:routes}
		\begin{tabular}{cc}
			\toprule %\noalign{\smallskip}
			Separated routes &  NME(\%) \\
			\midrule
			$K=2$ & $1.62$ \\
			$K=3$ & $\mathbf{1.55}$ \\
			$K=4$ & $1.63$ \\
			$K=5$ & $1.66$ \\
			$K=6$ & $1.67$ \\
			$K=7$ & $1.66$ \\
			\bottomrule
		\end{tabular}
	\end{minipage}
	\vspace{-1em}
\end{table}
\begin{table}[t]
	\centering
	\scriptsize
	\caption{A comparison of different $G$(left) and $T$(right) configurations on the AFLW dataset using the Vanilla CNN network.}
	\begin{minipage}{0.4\linewidth}
		% 		\centering
		\label{table:groups}
		\begin{tabular}{cc}
			\toprule
			Groups number &  NME(\%) \\
			\midrule
			$G = 1$ & $1.59$ \\
			$G = 2$ & $\mathbf{1.55}$ \\
% 			$G = 4$ & $1.55$ \\
			$G = 8$ & $1.57$ \\
% 			$G = 16$ & $1.55$ \\
% 			$G = 32$ & $1.58$ \\
			$G = 64$ & $1.55$ \\
			\bottomrule
		\end{tabular}
	\end{minipage}
	\hspace{1em}
	\begin{minipage}{0.4\linewidth}
		% 		\centering
		\label{table:poolingsize}
		\begin{tabular}{cc}
			\toprule %\noalign{\smallskip}
			Pooling size &  NME(\%) \\
			\midrule
			$T=1$ & $1.67$ \\
			$T=2$ & $1.58$ \\
			$T=3$ & $\mathbf{1.55}$ \\
			$T=4$ & $1.64$ \\
			\bottomrule
		\end{tabular}
	\end{minipage}
	\vspace{-1em}
\end{table}
\begin{table}[!t]
	%\small
	\footnotesize
	\begin{center}
		\caption{A study of the impact of the SepBN modules at different input resolutions in Vanilla CNN. $\checkmark$ indicates replacing BN with SepBN, -- indicates reserving the original standard BN.}
		\label{table:dyanmic-layers}
		\begin{tabular}{c c c c c c|c}
			\specialrule{.1em}{.05em}{.05em} 
			\multicolumn{6}{c|}{Input Resolution} & WFLW-test\\
			$64^2$ & $32^2$ & $16^2$ & $8^2$ & $4^2 $& $2^2 $&NME(\%)   \\
			\hline
			% 			--	&--	& 	--	& 	--	& 	--	& \checkmark & $5.76$ 	\\
			--	&--	& 	--	& 	--	& \checkmark & \checkmark &  $5.70$  \\
			% 			--	&--	& --	&\checkmark & \checkmark & \checkmark & $5.64$	\\   
			--	&--	&\checkmark&\checkmark & \checkmark &  \checkmark & $5.66$	\\   
			% 			--	&\checkmark&\checkmark&\checkmark & \checkmark &  \checkmark & $5.57$ \\   
			\checkmark	&\checkmark&\checkmark&\checkmark & \checkmark & \checkmark  & $\mathbf{5.48}$	\\   
			% 			\checkmark	&\checkmark&\checkmark&\checkmark & \checkmark & --  & $5.72$ \\   
			\checkmark	&\checkmark&\checkmark&\checkmark &--  & --  & $5.77$\\   
			% 			\checkmark	&\checkmark&\checkmark&-- & -- & --  & $5.79$	\\   
			\checkmark	&\checkmark&--&-- & -- & --  & $5.75$	\\   
			% 			\checkmark	&--&--&-- & -- & --  & $5.79$	\\ 
			--	&--&--&-- & -- & --  & $5.70$	\\  
			\specialrule{.1em}{.05em}{.05em}
		\end{tabular}
	\end{center}
% 	\vspace{-1.5em}
\end{table}

\subsection{Comparison with the state of the art}
\paragraph{COFW} We first compare our method with the state-of-the-art algorithms on COFW using NME and and failure rate in Table~\ref{table:comparison_cofw_testset}. `Vanilla CNN~(BN)' is the baseline network. `Vanilla CNN~(CNT)' uses the `Vanilla CNN~(BN)' as the shared backbone prototype. It is clear that CNT boosts the performance of the network considerably, validating the proposed cross-protocal network training strategy. `Vanilla CNN~(SepBN)' replaces all the BN layers in the baseline network with SepBN modules and improves the performance as well. `Vanilla CNN~(CNT+SepBN)' uses `Vanilla CNN~(SepBN)' as the shared backbone prototype to apply CNT and the result outperforms all the other settings as well as the state-of-the-art methods.

\renewcommand{\arraystretch}{1.1}
\begin{table}[t]
	\setlength{\tabcolsep}{12.5pt}
	\scriptsize
	\centering
	\caption{A comparison with the state-of-the-art methods on COFW in terms of NME and failure rate. Lower is better.}
	\label{table:comparison_cofw_testset}
	\begin{tabular}{lcc}
		\toprule
		&  NME(\%) &  Failure Rate(\%) \\
		\midrule
		HPM \cite{zhu2016unconstrained}&$7.50$ & $13$\\
		RAR \cite{xiao2016robust}&$6.03$ & $4.14$ \\
		DAC-CSR \cite{feng2017dynamic}&$6.03$ & $4.73$\\
		RSR \cite{cui2018recurrent}& $5.63$ & -\\
		LAB \cite{wu2018look}& $3.92$ & $0.39$\\
		ODN \cite{zhu2019robust}& $5.30$ & -\\
		RWing \cite{feng2019rectified} & $4.80$ & $3.16$ \\
		\hline
		Vanilla CNN (BN) & $4.07$ & $0.59$ \\
		Vanilla CNN (CNT) & $3.84$ & $0.20$ \\
		Vanilla CNN (SepBN) & $4.03$ & $0.39$ \\
		Vanilla CNN (CNT+SepBN) & $\mathbf{3.63}$ & $\mathbf{0.00}$ \\
		\bottomrule
	\end{tabular}
	\vspace{-1em}
\end{table}

\paragraph{AFLW} Table~\ref{table:comparison_aflw_testset} reports the evaluation results obtained on the AFLW dataset. The slight decline in performance of `Vanilla CNN (CNT)' on the AFLW-Frontal suggests that the diversified feature extraction capability of the backbone does not benefit a relatively simple test set. `Vanilla CNN~(SepBN)' achieves the state of the art. The performance of `Vanilla CNN~(CNT+SepBN)' drops slightly, but the performance of `Vanilla CNN~(CNT)' is boosted, verifying the merits of the proposed SepBN method.

\renewcommand{\arraystretch}{1.1}
\begin{table}[!t]
	\setlength{\tabcolsep}{12.5pt}
	\scriptsize
	\centering
	\caption{Facial landmark localization results (NME) on AFLW-Full and AFLW-Frontal. Lower is better.}
	\label{table:comparison_aflw_testset}
	\begin{tabular}{lcc }
		\toprule
		&  AFLW-Full &  AFLW-Frontal \\
		\midrule
		DAC-CSR \cite{feng2017dynamic}&$2.27$ & $1.81$\\
		TSR \cite{lv2017deep}&$2.17$ & - \\
		CPM + SBR \cite{dong2018supervision}&$2.14$&-\\
		SAN \cite{dong2018style}&$1.91$ & $1.85$\\
		DSRN \cite{miao2018direct}&$1.86$ & -\\
		LAB \cite{wu2018look}& $1.85$ & $1.62$\\
		ODN \cite{zhu2019robust}& $1.63$ & $1.38$\\
		SA \cite{liu2019semantic}& $1.60$ & - \\
		LUVLi \cite{kumar2020luvli}& $2.30$ & - \\
		3FabRec \cite{browatzki20203fabrec}& $1.84$ & $1.59$\\
		\hline
		Vanilla CNN (BN) & $1.65$ & $1.40$ \\
		Vanilla CNN (CNT) & $1.61$ & $1.42$ \\
		Vanilla CNN (SepBN) & $\mathbf{1.55}$ & $1.39$ \\
		Vanilla CNN (CNT+SepBN) & $1.57$ & $\mathbf{1.37}$ \\
		\bottomrule
	\end{tabular}
	\vspace{-1em}
\end{table}

\paragraph{WFLW} The evaluation performance on different WFLW testsets in different configurations is shown in Table~\ref{table:comparison_wflw_testset}. Note that when CNT is applied to the WFLW dataset~(`Vanilla CNN~(CNT)'), the performance improvement is much higher than that on the AFLW dataset. This is mainly because the size of WFLW training set is much smaller than the size of AFLW (20,000 vs 7,500), which makes the performance improvement of CNT on WFLW more significant. `Vanilla CNN~(SepBN)' easily surpasses the baseline network and `Vanilla CNN~(CNT+SepBN)' outperforms all the other algorithms in accuracy. 

\renewcommand{\arraystretch}{1.1}
\begin{table}[!t]
	\scriptsize
	%\tiny
	\setlength{\tabcolsep}{1.0pt}
	\centering
	\caption{Facial landmark localization results (NME) on WFLW \texttt{test} and $6$ subsets: pose, expression (expr.), illumination (illu.), make-up (mu.), occlusion (occu.)	and blur. Lower is better.}
	\label{table:comparison_wflw_testset}
	\begin{tabular}{lccccccc}
		\toprule
		& test & pose & expr. & illu. & mu. & occu. & blur\\
		
		\hline
		ESR \cite{cao2014face}& $11.13$ & $25.88$ & $11.47$ & $10.49$ & $11.05$ & $13.75$ & $12.20$\\
		SDM \cite{xiong2013supervised}& $10.29$ & $24.10$ & $11.45$ & $9.32$ & $9.38$ & $13.03$ & $11.28$\\
		CFSS \cite{zhu2015face}& $9.07$ & $21.36$ & $10.09$ & $8.30$ & $8.74$ & $11.76$ & $9.96$\\
		DVLN \cite{wu2017leveraging} & $6.08$ & $11.54$ & $6.78$ & $5.73$ & $5.98$ & $7.33$ & $6.88$\\
		LAB \cite{wu2018look} & $5.27$ & $10.24$ & $\mathbf{5.51}$ & $5.23$ & $\mathbf{5.15}$ & $6.79$ & $6.32$ \\
		RWing \cite{feng2019rectified} & $5.60$ & $9.79$ & $6.16$ & $5.54$ & $6.65$ & $7.05$ & $6.41$\\
		3FabRec \cite{browatzki20203fabrec} & $5.62$ & $10.23$ & $6.09$ & $5.55$ & $5.68$ & $6.92$ & $6.38$\\
		\hline
		Vanilla CNN (BN) & $5.70$ & $10.56$ & $6.15$ & $5.50$ & $5.65$ & $7.27$ & $6.47$\\
		Vanilla CNN (CNT) & $5.33$ & $9.46$ & $5.91$ & $5.17$ & $5.20$ & $6.55$ & $6.09$\\
		Vanilla CNN (SepBN) & $5.48$ & $9.97$ & $5.89$ & $5.36$ & $5.55$ & $6.83$ & $6.25$\\
		Vanilla CNN (CNT+SepBN) & $\mathbf{5.26}$ & $\mathbf{9.33}$ & $5.79$ & $\mathbf{5.09}$ & $5.20$ & $\mathbf{6.43}$ & $\mathbf{6.03}$\\
		\bottomrule
	\end{tabular}
	\vspace{-1em}
\end{table}
\begin{figure}[t]
	\centering
	\includegraphics[width=0.8\linewidth]{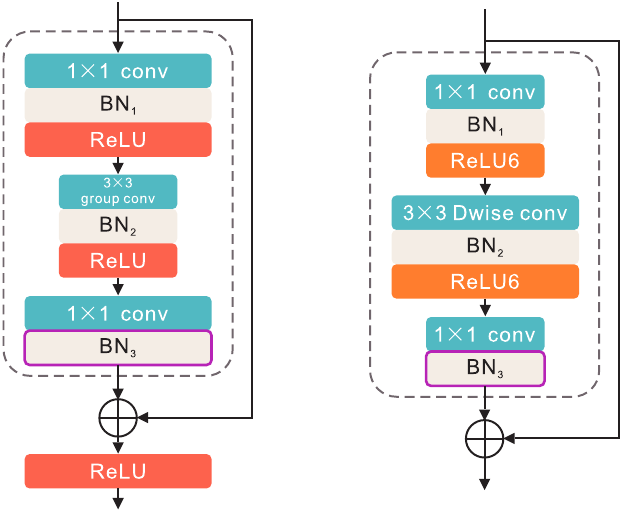}
	\caption{The corresponding Bottleneck Block used in ResNeXt-50 (left) and MobileNetV2 (right), the experiments results shows that replacing the $BN_3$ with the proposed SepBN module can maximise the performance.}
	\label{fig:bottleneck}
	\vspace{-1em}
\end{figure}

\subsection{Application to Modern Network architectures}
To further demonstrate the effectiveness of the proposed method for advanced network architectures, we apply SepBN and CNT to ResNeXt-50~\cite{xie2017aggregated} and MobileNetV2~\cite{sandler2018mobilenetv2}. 
For each of them, the parameters pre-trained on the ImageNet dataset are kept and the last linear layer is changed by setting the output length to $2L$. Note that the input image size is still $128 \times 128 \times 3$ for these two networks for the purpose of fair comparison.

For the usage of SepBN module in ResNeXt-50 and MobileNetV2, experiments were conducted to find the best solution to combine Bottleneck Blocks which are shown in Figure~\ref{fig:bottleneck} and the proposed SepBN module. As from Table~\ref{table:bottleneckbnconfiguremobile} tells, by replacing the original $BN_3$ in the Bottelneck Block with our proposed SepBN module, both ResNeXt-50 and MobileNetV2 shows the largest performance boost compared with the baseline network. 

\begin{table}[!t]
% 	\centering
	\scriptsize
	\caption{A study of the best combination solution of SepBN module and Bottleneck Block validated on COFW with ResNeXt-50 (left) and MobileNetV2 (right). $\checkmark$ indicates replacing BN with SepBN, -- indicates reserving the original standard BN, NaN indicates the network shows no convergence.}
	\begin{minipage}{0.4\linewidth}
		\label{table:bottleneckbnconfigureres}
		\begin{tabular}{c c c|c}
			\specialrule{.1em}{.05em}{.05em} 
			\multicolumn{3}{c|}{Different locations} & COFW\\
			$BN_{1}$ & $BN_{2}$ & $BN_{3}$ & NME(\%)   \\
			\hline
			-- & -- & -- & $3.95$  \\ 
			\checkmark & -- & -- & $4.03$  \\ 
			-- & \checkmark & -- & $4.12$  \\ 
			-- & -- & \checkmark & $\mathbf{3.51}$  \\ 
			\checkmark & \checkmark & -- & $4.03$  \\ 
			\checkmark & -- & \checkmark & $3.80$  \\ 
			-- & \checkmark & \checkmark & NaN  \\ 
			\checkmark & \checkmark & \checkmark & NaN  \\ 
			\specialrule{.1em}{.05em}{.05em}
		\end{tabular}
	\end{minipage}
	\hspace{2.9em}
	\begin{minipage}{0.4\linewidth}
		\label{table:bottleneckbnconfiguremobile}
		\begin{tabular}{c c c|c}
			\specialrule{.1em}{.05em}{.05em} 
			\multicolumn{3}{c|}{Different locations} & COFW\\
			$BN_{1}$ & $BN_{2}$ & $BN_{3}$ & NME(\%)   \\
			\hline
			-- & -- & -- & $5.07$  \\ 
			\checkmark & -- & -- & $4.07$  \\ 
			-- & \checkmark & -- & $4.03$  \\ 
			-- & -- & \checkmark & $\mathbf{3.96}$  \\ 
			\checkmark & \checkmark & -- & $4.08$  \\ 
			\checkmark & -- & \checkmark & $4.02$  \\ 
			-- & \checkmark & \checkmark & $4.10$  \\ 
			\checkmark & \checkmark & \checkmark & $4.05$  \\ 
			\specialrule{.1em}{.05em}{.05em}
		\end{tabular}
	\end{minipage}
	\vspace{-1em}
\end{table}

\renewcommand{\arraystretch}{1.1}
\begin{table}
	\scriptsize
	\centering
	\caption{A comparison of facial landmark localization results (NME) on COFW, AFLW-Full and WFLW using different configurations under different architectures. $\textcolor{red}{\downarrow}$ and $\textcolor{blue}{\uparrow}$ indicate the change of NME compared with the corresponding network architecture baseline. }
	\label{table:comparison_aflw_modern}
	\begin{tabular}{l|lll}
		\toprule 
		&  COFW & WFLW-test  & AFLW-Full \\
		\midrule
		Vanilla CNN (BN) & $4.07$ & $5.70$ & $1.65$ \\
		Vanilla CNN (CNT) & $3.84_{\textcolor{red}{\downarrow 0.23}}$ & $5.33_{\textcolor{red}{\downarrow 0.37}}$ & $1.61_{\textcolor{red}{\downarrow 0.04}}$ \\
		Vanilla CNN (SepBN) & $4.03_{\textcolor{red}{\downarrow 0.04}}$ & $5.48_{\textcolor{red}{\downarrow 0.22}}$ & $1.55_{\textcolor{red}{\downarrow 0.10}}$ \\
		Vanilla CNN (CNT+SepBN) & $3.63_{\textcolor{red}{\downarrow 0.44}}$ & $5.26_{\textcolor{red}{\downarrow 0.44}}$ & $1.57_{\textcolor{red}{\downarrow 0.08}}$ \\
		\hline
		MobileNetV2 (BN) & $5.07$ & $6.18$ & $1.66$ \\
		MobileNetV2 (CNT) & $3.57_{\textcolor{red}{\downarrow 1.50}}$ & $5.18_{\textcolor{red}{\downarrow 1.00}}$ & $1.64_{\textcolor{red}{\downarrow 0.02}}$ \\
		MobileNetV2 (SepBN) & $3.96_{\textcolor{red}{\downarrow 1.11}}$ & $5.31_{\textcolor{red}{\downarrow 0.87}}$ & $1.57_{\textcolor{red}{\downarrow 0.09}}$ \\
		MobileNetV2 (CNT+SepBN) & $3.56_{\textcolor{red}{\downarrow 1.51}}$ & $5.13_{\textcolor{red}{\downarrow 1.05}}$ & $1.60_{\textcolor{red}{\downarrow 0.06}}$ \\
		\hline
		ResNeXt-50 (BN) & $3.95$ & $4.90$ & $1.50$ \\
		ResNeXt-50 (CNT) & $3.49_{\textcolor{red}{\downarrow 0.46}}$ & $4.88_{\textcolor{red}{\downarrow 0.02}}$ & $1.48_{\textcolor{red}{\downarrow 0.02}}$ \\
		ResNeXt-50 (SepBN) & $3.51_{\textcolor{red}{\downarrow 0.44}}$ & $4.85_{\textcolor{red}{\downarrow 0.05}}$ & $1.46_{\textcolor{red}{\downarrow 0.04}}$ \\
		ResNeXt-50 (CNT+SepBN) & $3.47_{\textcolor{red}{\downarrow 0.48}}$ & $4.87_{\textcolor{red}{\downarrow 0.03}}$ & $1.48_{\textcolor{red}{\downarrow 0.02}}$ \\
		\bottomrule
	\end{tabular}
	\vspace{-1em}
\end{table}

We evaluate our Vanilla CNN, ResNeXt-50 and MobileNetV2 on the COFW, AFLW and WFLW datasets. The results are reported in Table~\ref{table:comparison_aflw_modern}.
`ResNeXt-50 (BN)' and `MobileNetV2 (BN)' are the baseline networks equipped with standard BN layers. SepBN modules with the same configuration as described above are used to replace the original $BN_3$ in all the Bottleneck Blocks, thus forming `ResNeXt-50 (SepBN)' and `MobileNetV2 (SepBN)'. `ResNeXt-50 (CNT)' and `MobileNetV2 (CNT)' are the networks evaluated using the CNT strategy. `ResNeXt-50 (CNT+SepBN)' and `MobileNetV2 (CNT+SepBN)' indicates the same training process as `Vanilla CNN (CNT+SepBN)'. 

According to the results, we can first see that the proposed SepBN module improves the performance of all the network architectures.
Second, the use of ResNeXt-50 further improves the final performance as compared with our Vanilla CNN.
However, MobileNetV2 performs slightly worse because of its lightweight network design.
Last, the use of the proposed CNT strategy can boost the performance of the trained network on small training dataset (COFW) significantly. However, for large capacity networks trained on relatively larger datasets, the use of the proposed CNT method may not able to improve the performance much.

%%% Conclusion
%%%------------------------------------------------------------
%%%------------------------------------------------------------
%%%------------------------------------------------------------
%%%------------------------------------------------------------
%%%------------------------------------------------------------
%%%------------------------------------------------------------
\section{Conclusion}
\label{sec_5}
In this paper, we presented a novel Separable Batch Normalization (SepBN) module for robust facial landmark localization. The SepBN module implicitly learns several sets of mapping parameters, corresponding to different cohorts of embedded features, to map the normalized features adaptively according to the attention weights produced by the attention branch. In addition, a training strategy called Cross-protocol Network Training~(CNT) was proposed and applied to enhance the network feature extraction capability by utilizing multiple datasets with different landmark definition schemes for network pre-training, and for fine-tuning on a specific dataset. Experiments on multiple networks with different architectures demonstrate the effectiveness and applicability of the proposed method.

\ifCLASSOPTIONcaptionsoff
  \newpage
\fi



\bibliographystyle{IEEEtran}
%\bibliography{IEEEabrv,../bib/ref}
\bibliography{ref.bib}
\end{document}